\documentclass[11pt]{article}

\usepackage{amstext}
\usepackage{epsfig}
\usepackage{amsthm}
\usepackage{latexsym}

\newtheorem{theorem}{Theorem}
\newtheorem{lemma}[theorem]{Lemma}

\newtheorem{fact}[theorem]{Fact}

\newcommand{\toe}{e}
\newcommand{\rmast}{\mbox{\sc mast}}
\newcommand{\mwm}{\mbox{\sc mwm}}

\newcommand{\sideT}{\mbox{\small \sc side-tree}}
\newcommand{\inp}{\mbox{\sc inp}}

\newcommand{\cntr}{{\rm hvy}}
\newcommand{\cpdecom}{{\cal D}}
\newcommand{\cpattach}{{\cal A}}

\newcommand{\secW}{\mbox{\sc s{\small ec}w}}

 \newcommand{\mam}{\mbox{\sc mam}}

\newcommand{\mgraph}{{\cal G}}
\newcommand{\maxsideR}{\mbox{\sc max}_R}
\newcommand{\maxsideL}{\mbox{\sc max}_L}

\begin{document}
\title{An Even Faster and More Unifying Algorithm
for Comparing Trees via Unbalanced Bipartite
Matchings\footnote{This paper combines results
of three conference papers:
(1) A faster and unifying algorithm for comparing trees. In
{\em Proceedings of the 11th Symposium on Combinatorial Pattern Matching},
pages 129-142, 2000.
(2) Unbalanced and hierarchical bipartite matchings with applications to
labeled tree comparison. In
{\em Proceedings of the 11th International
Symposium on Algorithms and Computation,} pages 479-490, 2000.
(3) A decomposition theorem for maximum weight bipartite matchings
with applications to evolutionary trees. In
{\em Proceedings of the 8th Annual European Symposium on Algorithms,}
pages 438-449, 1999.}}
\author{Ming-Yang Kao\thanks{Department of Computer Science,
Yale University, New Haven,
CT 06520, USA; kao-ming-yang@cs.yale.edu.
Research supported in part by NSF Grant CCR-9531028.}
\and
Tak-Wah Lam\thanks{Department of Computer Science and Information
Systems, University of Hong Kong, Hong Kong;
\{twlam, hfting\}@cs.hku.hk.
Research supported in part by Hong Kong RGC Grant HKU-7027/98E.}
\and
Wing-Kin Sung\thanks{E-Business Technology Institute,
University of Hong Kong, Hong Kong;
wksung@eti.hku.hk.}
\and
Hing-Fung Ting\footnotemark[3]}
\maketitle

\begin{abstract}
A widely used method for determining
the similarity of two labeled trees is to
compute a maximum agreement subtree of the two trees.
Previous work on this similarity measure is
only concerned with the comparison of labeled trees of two special kinds,
namely, {\it uniformly labeled} trees (i.e., trees with
all their nodes labeled by the same symbol) and {\it evolutionary} trees
(i.e., leaf-labeled trees with distinct symbols for distinct leaves).
This paper presents an algorithm for comparing trees that
are labeled in an arbitrary manner.
In addition to this generality, this algorithm is faster than
the previous algorithms.

Another contribution of this paper is on maximum weight bipartite
matchings.  We show how to speed up the best known matching algorithms when
the input graphs are node-unbalanced or weight-unbalanced.  Based on
these enhancements, we obtain an efficient algorithm for a
new matching problem called the {\it hierarchical bipartite matching} problem,
which is at the core of our maximum agreement subtree
algorithm.
\end{abstract}

\section{Introduction} \label{sec:introduction}

A {\it labeled} tree is a rooted tree
with an arbitrary subset of nodes labeled with symbols.
In recent years, many algorithms for comparing such trees
have been developed
for diverse application areas including
biology \cite{Finden:1985:OCP,Le:1989:RSS,Shapiro:1990:CMR},
chemistry \cite{Takahashi:1987:RLC},
linguistics \cite{Friedman:1981:ELF,Materna:1985:LCT},
computer vision \cite{Kimia:1995:SSD},
and structured text databases \cite{Kilpelainen:1991:RHT,Kilpelainen:1992:GTM,Mannila:1990:QLP}.

A widely used measure of the similarity of two labeled trees
is the notion of a maximum agreement subtree defined as follows.
A labeled tree $R$ is
a {\em label-preserving homeomorphic} subtree of another labeled tree $T$
if there exists a one-to-one mapping $f$ from the nodes of $R$ to
those of $T$ such that for any nodes $u, v, w$ of $R$,
 (1) $u$ and $f(u)$ have the same label; and
 (2) $w$ is the least common ancestor of $u$ and $v$ if and only if $f(w)$ is
the least common ancestor of $f(u)$ and $f(v)$.
Let $T_1$ and $T_2$ be two labeled trees.
An {\em agreement} subtree of $T_1$ and $T_2$
is a labeled tree which is also a label-preserving homeomorphic subtree of
the two trees.
A {\em maximum} agreement subtree
is one which maximizes the number of labeled nodes.
Let \rmast$(T_1,T_2)$ denote the number
of labeled nodes in a maximum agreement subtree of $T_1$ and $T_2$.

In the literature, many algorithms for computing
a maximum agreement subtree
have been developed.
These algorithms focus on
the special cases where
$T_1$ and $T_2$ are
either (1) {\it uniformly labeled} trees,
i.e., trees with all their nodes unlabeled, or equivalently,
labeled with the same symbol
or
(2) {\it evolutionary trees} \cite{Hillis:1996:MS}, i.e.,
leaf-labeled trees with distinct symbols for distinct leaves.

We denote $n$
as the number of nodes in the labeled trees $T_1$ and $T_2$,
and $d$ as the maximum degree of $T_1$ and $T_2$.
For uniformly labeled trees $T_1$ and $T_2$,
Chung \cite{Chung:1987:TAS} gave an algorithm to determine whether $T_1$ is
a label-preserving homeomorphic subtree of $T_2$
using $O(n^{2.5})$ time.
Gupta and Nishimura \cite{Gupta:1998:FLS}
gave an algorithm which actually computes
a maximum agreement subtree of $T_1$ and $T_2$
in $O(n^{2.5} \log n)$ time.
For evolutionary trees, Steel and Warnow \cite{Steel:1993:KTT}
gave the first polynomial-time algorithm
for computing a maximum agreement subtree.
Farach and Thorup \cite{Farach:1997:SDP} improved the time
complexity from  $O(n^{4.5}\log{n})$ to $O(n^{1.5}\log n)$.
Faster algorithms for the case $d = O(1)$
were also discovered. The algorithm of Farach,
Przytycka and Thorup \cite{Farach:1995:CAT} runs in $O(\sqrt{d} n \log^3 n)$ time,
and that of Kao \cite{Kao:1998:TCE} takes $O(n d^2 \log^2 n \log d)$ time.
Cole {\em et al.}~\cite{Cole:2000:AMA}
gave an $O(n \log n)$-time algorithm for
the case where $T_1$ and $T_2$ are binary trees.
Przytycka \cite{Przytycka:1997:SDP} attempted to generalize
the algorithm of Cole {\em et al.}~so
that the degree-2 restriction could be removed
with the running time being $O(\sqrt{d} n \log n)$.

For {\em unrestricted} labeled trees
(i.e., trees where
labels are not restricted to leaves and may not be distinct),
little work has been reported,
but they have applications in several contexts
\cite{Abiteboul:1999:VX,Kimia:1995:SSD}.
For example, labeled trees are used to represent
sentences in a structural text database \cite{Kilpelainen:1991:RHT,Kilpelainen:1992:GTM,Mannila:1990:QLP}
and querying such a database
involves comparison of trees; an XML document can also be represented
by a labeled tree
\cite{Abiteboul:1999:VX}.
Instead of solving special cases,
this paper gives an algorithm to compute
$\rmast(T_1, T_2)$ where $T_1$ and $T_2$
are unrestricted labeled trees. As detailed below,
our algorithm not only is more general
but also uniformly improves or matches the previously best algorithms
for subtree homeomorphism and evolutionary tree comparison.

\newcommand{\M}{\Delta}

Let $\M_{T_1, T_2}$
(or simply $\M$ when the context is clear)
$=$ $\sum_{u \in T_1} \sum_{v \in T_2} \delta(u, v)$
where $\delta(u, v) = 1$ if nodes $u$ and $v$
are labeled with the same symbol, and $0$ otherwise.
Our algorithm computes $\rmast(T_1, T_2)$ in
$O(\sqrt{d} \M \log \frac{2n}{d})$ time.
Thus, if $T_1$ and $T_2$ are uniformly labeled trees,
then $\M  \leq n^2$ and the time complexity
of our algorithm is $O(\sqrt{d} n^2 \log \frac{2n}{d})$,
which is faster than the Gupta-Nishimura
algorithm \cite{Gupta:1998:FLS} for any $d$.
If $T_1$ and $T_2$ are evolutionary trees,
then $\M \leq n$ and the time complexity of our algorithm
is $O(\sqrt{d} n \log \frac{2n}{d})$,
which is better than the
$O(\sqrt{d} n \log n)$ bound claimed by Przytycka \cite{Przytycka:1997:SDP}.
In particular, our algorithm can
attain the $O( n \log n)$ bound for binary trees \cite{Cole:2000:AMA}.
Also for general evolutionary trees,
our algorithm runs in $O(n^{1.5})$ time since
$\sqrt{d} n \log \frac{2n}{d} = O(n^{1.5})$ for any degree $d$.
This is faster than the $O(n^{1.5} \log n)$ time of the
Farach-Thorup algorithm \cite{Farach:1997:SDP}.

The efficiency achieved by our \rmast\ algorithm is based on
improved algorithms for computing maximum weight matchings of bipartite
graphs that satisfy some structural properties.
Let $G=(X,Y,E)$ be a bipartite graph with positive integer weights
on its edges.  Denote by $n$, $m$, $N$, and $W$ the number of nodes,
the number of edges, the maximum edge weight, and the total
edge weight of $G$, respectively.
The best known algorithm for computing maximum weight bipartite
matchings was given by Gabow and Tarjan \cite{Gabow:1989:FSA}, which takes
$O(\sqrt{n}m \log nN)$ time.
For some applications where the total edge weight is small (say, $W = O(m)$),
Kao {\em et al.}~\cite{Kao:2000:DTM} gave a slightly faster
algorithm that runs in $O(\sqrt{n}W)$ time.
Intuitively, a bipartite graph is
{\em node-unbalanced} if
there are much fewer nodes on one side than the other.
It is {\em weight-unbalanced} if its total weight is dominated by the edges
incident to a few nodes; we call these nodes the {\it dominating} nodes.
In this paper, we show how to enhance these two matching algorithms
when the input graphs are either {node-unbalanced} or
{weight-unbalanced}.

The node-unbalanced property has many
practical applications (see, e.g., \cite{Gusfield:1987:FAB})
and has been exploited to
improve various graph algorithms. For example,
Ahuja {\it et al.}~\cite{Ahuja:1994:IAB} adapted
several bipartite network flow algorithms
such that the running times depend on the number
of nodes in the smaller side of the input bipartite graph
instead of the total number of nodes.
Tokuyama and Nakano used this property to
reduce the time complexity of
the minimum cost assignment problem \cite{Tokuyama:1995:GAM} and
the Hitchcock transportation problem \cite{Tokuyama:1995:EAH}.
This paper presents similar improvements for maximum weight matching.
Specifically, we show that the running time
of the matching algorithms of Gabow and Tarjan  \cite{Gabow:1989:FSA} and
Kao {\em et al.}~\cite{Kao:2000:DTM} can be improved to
$O(\min\{\sqrt{n_s}m\log n_s N, m+n_s^{2.5}\log n_s N\})$
and $O(\sqrt{n_s}W)$, respectively, where
$n_s$ is the number of nodes in the smaller side of the
input bipartite graph.

The weight-unbalanced property is exploited in another way.
Given a weight-unbalanced bipartite graph $G$,
let $G'$ be the subgraph of $G$ with its dominating nodes removed.
Note that $G'$ has a total weight much smaller than $G$ does.
Based on the $O(\sqrt{n}W)$-time matching algorithm of Kao
{\it et al.}~\cite{Kao:2000:DTM},
finding the maximum weight matching of $G'$ is much faster than
finding one of $G$.
To take advantage of this fact,
we design an efficient algorithm that finds a maximum
weight matching of $G$ from that of $G'$.
This algorithm is substantially faster than
applying directly the $O(\sqrt{n}W)$-time matching algorithm on $G$.

\begin{figure}[t]
\begin{center}
\epsfig{file=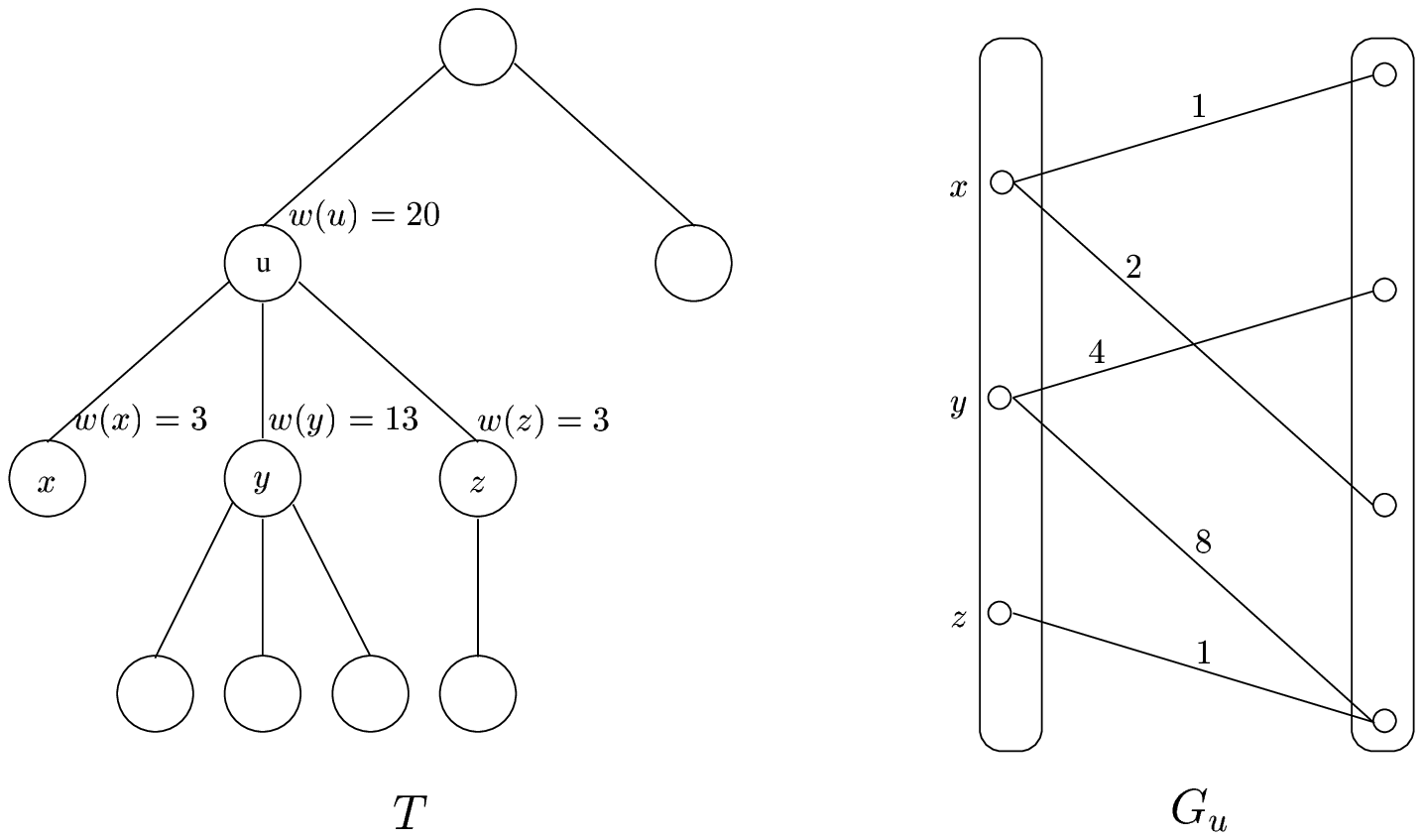, width=\textwidth}
\end{center}
\caption{$T$ is an instance of the hierarchical bipartite matching problem.
$G_u$ is the bipartite graph associated with the node $u$.}
\label{fig-hbm}
\end{figure}

These results for unbalanced graphs provide a basis for
solving a new matching problem
called the {\em hierarchical bipartite matching} problem.
This matching problem is at the core of our \rmast\ algorithm
and is defined as follows.
Let $T$ be a rooted tree.  Denote $r$ as the root of $T$.
Let $C(u)$ denote the set of children of node $u$.
Every node $u$ of $T$ is associated with a positive integer
$w(u)$ and a weighted bipartite graph $G_u$
satisfying the following properties:

\begin{itemize}
\item
        $w(u) \geq \sum_{v \in C(u)} w(v)$.
\item
        $G_u = (X_u,Y_u,E_u)$ where $X_u = C(u)$.
        Each edge of $G_u$ has a positive integer weight,
        and there is no isolated node.  For any node $v \in X_u$,
        the total weight of all the edges incident
        to $v$ is at most $w(v)$.
        Thus, the total weight of the edges in $G_u$ is at most $w(u)$.
\end{itemize}
See Figure~\ref{fig-hbm} for an example.
For any weighted bipartite graph $G$,
let $\mwm(G)$ denote
a maximum weight matching of $G$.
The {\em hierarchical matching problem} is to compute $\mwm(G_u)$
for all internal nodes $u$ of $T$.
Let $b = \max_{u \in T}\{ \min\{|X_u|,|Y_u|\}\}$ and
$e = \sum_{u\in T} |E_u|$.
The problem can be solved by applying directly our results for
node-unbalanced graphs; for example, it can be solved in
$O(\sum_{u \in T}\sqrt{b}|E_u|\log b w(u))
=O (\sqrt{b} e \log w(r))$ time using the
enhanced Gabow-Tarjan algorithm.
However, this time complexity is not yet satisfactory.
When comparing labeled trees,
we often encounter instances of the hierarchical bipartite matching problem
with $e$ being very large; in particular,
$e$ is asymptotically much greater than $w(r)$.
We further improve the running time
to $O(\sqrt{b}w(r) + e)$ by making additional use of our
technique for weight-unbalanced graphs and
exploiting trade-offs between
the size of the bipartite graphs involved and their total edge weight.

The rest of the paper is organized as follows.
Section~\ref{sec:unbalanced} details our techniques of
speeding up the existing algorithms for unbalanced graphs.
Section~\ref{sec:hi-matching} gives an efficient algorithm
for solving the hierarchical matching problem.  Finally,
Section~\ref{sec:mast} describes our algorithm for
computing maximum agreement subtrees.

\section{Maximum weight matching of unbalanced graphs}
\label{sec:unbalanced}
Throughout this section,
let $G=(X,Y,E)$ be a weighted bipartite graph with no isolated nodes.
Let $n= |X|+|Y|$, $n_s = \min \{ |X|, |Y|\}$,
$m=|E|$,  $N$ be the largest edge weight, and $W$ be the total edge weight.

Suppose every edge of $G$ has a positive integer weight.
Gabow and Tarjan \cite{Gabow:1989:FSA} and Kao {\em et al.}~\cite{Kao:2000:DTM}
gave an $O(\sqrt{n}m \log (n N))$-time algorithm and an $O(\sqrt{n}W)$-time
one to compute $\mwm(G)$, respectively.

\subsection{Matchings of node-unbalanced graphs}
The following theorem speeds up the computation of
$\mwm(G)$ if $G$ is node-unbalanced.
\begin{theorem}\label{thm:node-unbalanced}\

\begin{enumerate}
\item
$\mwm(G)$ can be computed in $O(\sqrt{n_s} m \log (n_sN))$ time.
\item
$\mwm(G)$ can be computed in
$O(m + n_s^{2.5}\log (n_sN))$ time.
\item
$\mwm(G)$ can be computed in $O(\sqrt{n_s} W)$ time.
\end{enumerate}
\end{theorem}
\begin{proof}
Without loss of generality,
we assume $n_s = |X| \leq |Y|$.
The statements are proved as follows.

Statement 1.
For any node $v$ in $G$, let $\alpha(v)$ be the number
of edges incident to $v$.
Suppose $Y = \{y_1, y_2,\ldots, y_{kn_s + r} \}$
where $k\geq 1$, $0 \leq r< n_s$, and
$\alpha(y_1) $ $ \leq $ $\alpha(y_2) \leq \cdots \leq \alpha(y_{kn_s+r})$.
We partition $Y$ into
$Y_0 = \{y_1, \ldots, y_r\},
Y_1 = \{y_{r+1}, \ldots, y_{r+n_s} \},
\ldots,$ and
$Y_k = \{y_{r+(k-1)n_s+1}, \ldots, y_{r+kn_s} \}$.
Note that except $Y_0$, every set has $n_s$ nodes.

For any $Y' \subseteq Y$, denote
$G(Y')$ as the subgraph of $G$ induced by
all the edges incident to $Y'$.
Suppose that $M_i$ is a maximum weight matching of
$G(Y_0 \cup Y_1 \cup \cdots \cup Y_i)$.
Let $Y_{M_i} = \{ y \mid (x, y) \in M_i \}$.
Note that a maximum weight matching
of $G(Y_{M_i} \cup Y_{i+1})$ is
also one of $G(Y_0 \cup Y_1
        \cup \cdots \cup Y_{i+1})$.
Therefore, we can compute $\mwm(G)$ using the following algorithm:
\begin{itemize}
\item
Step 1. Compute a maximum weight matching $M_0$ of $G(Y_0)$.
\item
Step 2. For $i = 1$ to $k$,
\newcounter{hhgt}
\begin{list}{\thehhgt}
{\usecounter{hhgt}\setcounter{hhgt}{0}\renewcommand{\thehhgt}{}
\setlength{\rightmargin}{0in}
\settowidth{\leftmargin}{Step 3.}\addtolength{\leftmargin}{\labelsep}}
\item
   let $Y_{M_{i-1}}$ = $\{ y \mid (x, y) \in M_{i-1} \}$;
\item
    compute a maximum weight matching $M_i$
    of $G(Y_{M_{i-1}} \cup Y_i)$.
\end{list}
\item
Step 3. Return $M_k$.
\end{itemize}

The running time is analyzed below.
Let $\alpha(Y')$ be the total number of edges in $G(Y')$.
For $1 \leq i \leq k$, $\alpha(Y_{M_{i-1}}) \leq \alpha(Y_i)$, and
$\alpha(Y_{M_{i-1}} \cup Y_i) \leq 2 \alpha(Y_i)$.
Using the matching algorithm by Gabow and Tarjan \cite{Gabow:1989:FSA},
we can compute $\mwm(G(Y_{M_{i-1}} \cup Y_{i}))$ in
$O(\sqrt{|Y_{M_{i-1}} \cup Y_i|} \alpha(Y_i)
\log (|Y_{M_{i-1}} \cup Y_i| N))$ time.
Note that
$|Y_{M_{i-1}}|\leq n_s$ and $|Y_i| = n_s$.
Hence, the whole algorithm uses
$O(\sum_{i=1}^k \sqrt{n_s} \alpha(Y_i) \log (n_s N))$
$=$ $O(\sqrt{n_s} m \log (n_s N))$ time.

Statement 2. Since we suppose $|X| \leq |Y|$,
any matching of $G$ contains at most $|X| = n_s$ edges.
Thus, for every $u \in X$, we can discard the edges
incident to $u$ that are not among the $n_s$ heaviest ones;
the remaining $n_s^2$ edges must still contain a maximum weight
matching of $G$.  Note that we can find these $n_s^2$
edges in $O(m)$ time, and from Statement 1, we can
compute $\mwm(G)$ from them in
$O(\sqrt{n_s} n_s^2 \log (n_s N))$ time.
The total time taken is
$O(m + n_s^{2.5}\log (n_sN))$.

Statement 3.
The algorithm in the proof of Statement 1
can be adapted to find $\mwm(G)$ in $O(\sqrt{n_s} W)$ time
by using the $O(\sqrt{n}W)$-time
matching algorithm of Kao {\em et al.}~\cite{Kao:2000:DTM}
to compute each $M_i$.
For any $Y' \subseteq Y$,
we redefine $\alpha(Y')$ to be the total weight of edges incident to $Y$.
Then we can use the same analysis to
show that the adapted algorithm runs
in $O(\sum_{i=1}^{k}\sqrt{n_s} \alpha(Y_i))$ =
$O(\sqrt{n_s} W)$ time.
\end{proof}

\subsection{Matchings of weight-unbalanced graphs}
We show how to speed up the matching algorithm of
Kao {\it et al.}~\cite{Kao:2000:DTM}
when the input graph $G$ is weight-unbalanced.
The key technique is stated in Lemma~\ref{lem-recover} below.
The following example illustrates how this lemma can help.
Suppose that $G$ has $O(1)$ dominating nodes.  Let
$G'$ be the subgraph of $G$ with the dominating nodes removed.
Let $W'$ be the total edge weight of $G'$. Since $G$ is
weight-unbalanced, we further assume $W' = o(W)$.
To compute $\mwm(G)$, we can first
use Theorem~\ref{thm:node-unbalanced}(3) to
compute $\mwm(G')$ in $O(\sqrt{n_s}W')$ time
and then use Lemma~\ref{lem-recover}
to compute $\mwm(G)$
from $\mwm(G')$ in $O(m\log n_s)$ time.
The total running time
is $O(m \log n_s + \sqrt{n_s}W') = o(\sqrt{n_s}W)$,
which is smaller than the running
time of using the
algorithm of Kao {\it et al.}~\cite{Kao:2000:DTM} to find $\mwm(G)$ directly.

\begin{lemma}\label{lem-recover}
Let $H = \{x_1, x_2, \ldots, x_h \}$ be a subset of $h$ nodes of $X$.
Let $G-H$ be the subgraph of $G$ constructed by removing the nodes in $H$.
Denote by $E'$ the set of edges in $G-H$.
Given $\mwm(G-H)$, we can compute
$\mwm(G)$ in $O(|E|+ (h^2|E'|+h^3)\log n_s)$ time.
\end{lemma}
\begin{proof}
First, we show that using $O(|E|)$ time,
we can find a set $\Upsilon$ of only $O(\min \{h |E'| + h^2, n_s^2\})$
edges such that $\Upsilon$ still contains
a maximum weight matching of $G$.
In the proof of Theorem~\ref{thm:node-unbalanced}(2),
it has already been shown
that we can find in $O(|E|)$ time a set
of $O(n_s^2)$ edges that contains $\mwm(G)$.
Thus, it suffices to find in $O(|E|)$ time
another set
of $O(h|E'|+h^2)$ edges that contains $\mwm(G)$;
$\Upsilon$ is just
the smaller of these two sets.
Let $Y'$ be the subset of nodes of $Y$ that are
endpoints of $E'$. For any $x_i \in H$, we select,
among the edges incident to $x_i$, a subset of edges
$E_i$,
which is the union of the following
two sets:
\begin{itemize}
 \item $\{ (x_i,y) \mid y \in Y' \}$;
 \item $\{ (x_i,y) \mid (x_i,y)$ is among the $h$ heaviest
        edges with  $y \not\in Y' \}$.
\end{itemize}
Observe that
 $E' \cup E_1 \cup \cdots \cup E_{h}$
must contain a maximum weight matching of $G$, and
these $|E' \cup E_1 \cup \cdots E_{h}| =
O(h|E'|+h^2)$ edges can be found in $O(|E|)$ time.

By discarding all unnecessary edges (i.e.,
edges neither in $\Upsilon$ nor in $\mwm(G-H)$),
we can assume that $G$ has only $O(\min\{h|E'|+h^2, n_s^2\})$ edges,
while still containing $\mwm(G)$ and $\mwm(G-H)$.
This preprocessing requires an extra $O(|E|)$
time for finding $\mwm(G)$.

Below, we describe a procedure which, given
any bipartite graph
$D$ and any node $x$ of $D$,
finds $\mwm(D)$ from $\mwm(D-\{x\})$ in
$O(m_D\log m_D)$ time, where $m_D$ is the number
edges of $D$.
Then, starting from $G-H$, we
can apply this procedure repeatedly
$h$ times to find
$\mwm(G)$ from $\mwm(G-H)$.
Since
$G$ is assumed to have only $O(\min\{h|E'|+h^2, n_s^2\})$ edges,
this process takes $O(h((h|E'|+h^2)\log n_s))$ time.
This lemma follows.

Let $M$ and $M_x$ be a maximum weight matching of $D$ and $D-\{x\}$,
respectively; denote by $S$ the set of augmenting paths and cycles
formed in $M \cup M_x - M \cap M_x$, and
let $\sigma$ be the augmenting path in $S$ starting from $x$.
 Note that the augmenting paths and cycles in $S - \{ \sigma \}$
 cannot improve the matching $M_x$; otherwise, $M_x$
 is not a maximum weight matching of $D -\{x\}$.
 Thus, we can transform $M_x$ to $M$
 using $\sigma$.
 Note that $\sigma$ is indeed a maximum augmenting path starting
 from $x$, which can be found in $O(m_D\log m_D)$ time \cite{Cormen:1990:IA}.
\end{proof}

\section{Hierarchical bipartite matching}
\label{sec:hi-matching}
Throughout this section,
let $T$ be a rooted tree as defined in the definition
of the hierarchical bipartite matching problem in
\S\ref{sec:introduction}.
The root of $T$ is denoted by $r$.
 For each node $u$ of $T$,
 $w(u)$ and $G_u=(X_u,Y_u,E_u)$ denote the weight and
 the bipartite graph associated with $u$, respectively.
 Furthermore, let
 $b = \max_{u \in T} \left \{ \min \{ |X_u|, |Y_u| \} \right \}$,
 and $\toe = \sum_{u \in T}|E_u|$.

 In this section, we describe an algorithm for computing
  $\mwm(G_u)$ for all $u \in T$ in $O(\sqrt{b}w(r)+\toe)$ time.
Our algorithm is based on two crucial observations.
One is that for any value $x$,
there are at most $w(r)/ x$ graphs with its second maximum
edge weight greater than $x$.
The other is that  most of these graphs have their total weight
dominated by edges incident to a few nodes.
For those graphs with a large second maximum edge weight,
we compute their maximum weight matchings
using a less weight-sensitive algorithm.
As there are not many of them,
the computation is efficient.
For the other graphs,
their weights are dominated by the
edges incident to a few nodes.  Thus, using
Lemma~\ref{lem-recover}
and a weight-efficient matching algorithm,
we can compute the maximum weight matchings for these
graphs efficiently.
Details are as follows.
 
 Consider any subset $B$ of nodes of $T$. Let $\delta = \min_{u \in B} w(u)$.
 We say that $B$ has a {\em critical degree}
 $h$ if for every $u \in B$, $u$ has at most $h$ children with
 weight at least $\delta$.
 For any internal node $u$, let
 $\secW(u)$ $=$ $\mbox{2nd-}\max\{ w(v) \mid v \in C(u)\}$, i.e.,
 the value of the second largest $w(v)$ over all the children $v$ of $u$.
 Lemma~\ref{lem-tradeoff} below shows
 the importance of  $\secW(u)$ and critical degrees.
 Lemma~\ref{lem-tradeoff}(\ref{item-tradeoff})
 shows that there are not many nodes $u$ with
 large $\secW(u)$; 
 for those nodes $u$ with small $\secW(u)$, they should not have
 a large critical degree, and Lemma~\ref{lem-tradeoff}(\ref{item-heavy})
 states that
 the maximum weight matchings associated with these nodes
 can be computed efficiently.

 \begin{lemma} \label{lem-tradeoff}\
 \begin{enumerate}
 \item \label{item-tradeoff}
 Let $x$ be any positive number.  Let $A$ be the
 set of nodes $u$ of $T$ with $\secW(u) > x$. Then
 $|A| < w(r) / x$.
 \item \label{item-heavy}
 Let $B$ be any set of nodes of $T$.  If $B$ has critical degree $h$,
 then we can compute $\mwm(G_u)$ for all $u \in B$ in
  $O((\sqrt{b} + h^3\log b)w(r)  + \sum_{u \in B} |E_u|)$  time.
 \end{enumerate}
 \end{lemma}
 \begin{proof}
 The statement are proved as follows.
\newcommand{\secC}{\mbox{\sc s{\small ec}C}}

Statement~\ref{item-tradeoff}.
Let $L$ be the set of nodes $u$ in $T$ such that
(1) $w(u) > x$
and (2) either $u$ is a leaf or
$w(v) \leq x$ for all children $v$ of $u$.
Since the subtrees rooted at the nodes of $L$
are disjoint, $w(r) \ge \sum_{u \in L} w(u) > x |L|$.
Thus, $|L| < w(r)/x$.
Let $T'$ be the tree in $T$ induced by $L$,
i.e., $T'$ contains exactly the nodes of $L$ and the least common
ancestor of every two nodes of $L$.
Note that $T'$ has at most $|L|$ leaves and
at most $|L|$ internal nodes.
On the other hand, every node of $A$ is an internal
node of $T'$; thus, $|A| \le |L| < w(r)/x$.

Statement~\ref{item-heavy}.
 For every node $u \in B$, let $H(u)$ be the set of
 $u$'s children that have a weight at least
  $\delta = \min_{u \in B} w(u)$ each. Let
 $L(u)$ be the set of the rest of $u$'s children.
 Note that $|H(u)| \leq h$ because $B$ has critical degree $h$.
 Since the weight of $G_u - H(u)$ is at most
 $\sum_{x \in L(u)} w(x)$ and $b \geq \min \{ |X_u|, |Y_u| \}$,
 by Theorem~\ref{thm:node-unbalanced}
 we can compute $\mwm(G_u - H(u))$
 in time
 \begin{equation} \label{eqn:g-u}
  \textstyle
   O(\sqrt{b}\sum_{x \in L(u)} w(x) + |E_u|).
 \end{equation}
 Since $G_u - H(u)$ has
 at most $\sum_{x \in L(u)}w(x)$ edges and $|H(u)| \leq h$,
 by Lemma~\ref{lem-recover}, we can compute $\mwm(G_u)$ from
 $\mwm(G_u - H(u))$ in time
 \begin{equation} \label{eqn:g}
 \textstyle
 O(|E_u| + (h^2\sum_{x \in L(u)}w(x) + h^3)\log b)
 = O(h^3 \log b \sum_{x \in L(u)} w(x) + |E_u|).
 \end{equation}
   From Equations (\ref{eqn:g-u}) and (\ref{eqn:g}), we can compute $\mwm(G_u)$
  for all $u \in B$ in time
  \[   \textstyle O\left (\sum_{u \in B}
        \Bigl( (\sqrt{b} + h^3 \log b)
    \sum_{x \in L(u)}w(x) + |E_u|\Bigr ) \right ).
  \]
 Since the subtrees rooted at some node in
 $\bigcup_{u \in B} L(u)$ are
 disjoint,
 $\sum_{u \in B}\sum_{x \in L(u)}w(x) \leq w(r)$.
 This statement follows.
 \end{proof}

We are now ready to compute $\mwm(G_u)$
for all nodes $u$ of $T$.  We divide all the nodes in $T$ into two
sets:  $ \Phi = \{ u \in T \mid \secW(u) > b^3 \}$ and
$\Pi = \{ u \in T \mid \secW(u) \leq b^3 \}$.

 Every node $u \in \Phi$ has $\secW(u) > b^3$;
 by Lemma~\ref{lem-tradeoff}(\ref{item-tradeoff}), $|\Phi| \leq w(r)/b^3$.
 Furthermore, by Theorem~\ref{thm:node-unbalanced}(2), the time for
 computing $\mwm(G_u)$ for all $u \in \Phi$ is
 $O(\sum_{u \in \Phi}
          ( b^{2.5} \log (b w(u)) + |E_u|))$ =
 $O(\frac{w(r)}{b^3}  b^{2.5} \log w(r) + \sum_{u \in \Phi}|E_u|)$ =
 $O(w(r) \frac{\log  w(r)}{\sqrt{b}} + \sum_{u \in \Phi}  |E_u|)$.
 This time complexity is still far from our goal as
 $\log w(r)$ may be much larger than $\sqrt{b}$.
 To improve the time complexity, we first note that
 using the technique for proving Lemma~\ref{lem-recover},
 we can compute $\mwm(G_u)$ in time depending only on $\secW(u)$.
 Then, with a better
 estimation of $\secW(u)$, we can reduce the time
 complexity to $O(w(r) + \sum_{u \in \Phi}|E_u|)$.
 Details are given in Lemma~\ref{lem:phi}(1).

 For $\Pi$, we can handle the nodes $u \in \Pi$ with
 $w(u) > b^3$ easily.  For nodes with $w(u) < b^3$, we apply
 Lemma~\ref{lem-tradeoff}(\ref{item-heavy})
 to compute $\mwm(G_u)$.
 The basic idea is to partition the nodes $u \in \Pi$ into a constant
 number of sets according to
 $w(u)$ such that every set has critical degree $b^{\frac{1}{7}}$.
 This can ensure that the total time to compute all the $\mwm(G_u)$ is
 $O(\sqrt{b}w(r) + \sum_{u \in \Pi}|E_u|)$.
 Details are given in Lemma~\ref{lem:phi}(2).

\begin{lemma} \label{lem:phi}\

\begin{enumerate}
\item
We can compute $\mwm(G_u)$ for
all $u \in \Phi$ in $O(w(r) +\sum_{u \in \Phi}|E_u|)$ time.
\item
We can compute $\mwm(G_u)$ for all $u \in \Pi$ in
   $O(\sqrt{b}w(r) + \sum_{u \in \Pi}|E_u|)$ time.
\end{enumerate}
\end{lemma}
\begin{proof}
The two statements are proved as follows.

Statement 1. Observe that for any $u \in \Phi$, $G_u$ has
at most $b^2$
edges relevant to the computation
of $\mwm(G_u)$, and they can be found in $O(|E_u|)$
time.
Let $E'_u$ be this set of edges.
Below, we assume that, for every $u \in \Phi$,
$G_u$ has only edges in $E'_u$.
Otherwise, it costs $O(\sum_{u \in \Phi}|E_u|)$ extra time
to find all $E_u'$ and the assumption holds.

For every $k \geq 1$, let
     $\Phi_k = \{ u \in \Phi \mid 2^{k-1} b^3 < \secW(u) \leq 2^{k}b^3 \}$.
Obviously, the nonempty sets $\Phi_k$ form a partition of $\Phi$.
Below, we show that for
any nonempty $\Phi_k$, we can compute $\mwm(G_u)$
for all $u \in \Phi_k$ in
$O\left (w(r) k/2^k+\textstyle{\sum_{u \in \Phi_k}|E'_u|}\log b\right)$
time.
Thus,
the time for computing $\mwm(G_u)$ for all $u \in \Phi$ is
$O(\sum_{k \geq 1} w(r) k/2^k + \sum_{u \in \Phi}|E'_u|\log b)$
$=$ $O(w(r) + \sum_{u \in \Phi}|E'_u|\log b)$ $=$
$O(w(r) + (w(r)/b^3)b^2 \log b )$ $=$ $O(w(r))$, and Statement 1 follows.

We now give the details of computing $\mwm(G_u)$ for all $u \in \Phi_k$.
Let $u'$ be the child of $u$ where $w(u')$ is the largest
over  all children of $u$.
Since $\secW(u) \leq 2^{k}b^3$,
every edge of $G_u - \{ u' \}$ has weight at most
$2^{k}b^3$.
By Theorem~\ref{thm:node-unbalanced}(2)
and Lemma~\ref{lem-recover}, and the fact $b \geq \min\{|X_u|,|Y_u|\}$,
we can find $\mwm(G_u)$ in
$O(\sqrt{b} b^2 \log (b 2^{k}b^3) + |E'_u|\log b)$ time.
By Lemma~\ref{lem-tradeoff}(\ref{item-tradeoff}),
$|\Phi_k| \leq \frac{ w(r)}{2^k b^3}$. Thus, we can compute
$\mwm(G_u)$ for all $u \in \Phi_k$ in time
\begin{eqnarray*}
\textstyle
    O\left (\sum_{u \in \Phi_k}\sqrt{b} b^2 \log (b 2^{k}b^3) + |E'_u| \log b\right)
    &=&
\textstyle
    O \left (\frac{w(r)b^{2.5}}{2^k b^3}(k+\log b) +
       \sum_{u \in \Phi_k} |E'_u|\log b \right) \\
    &=&
\textstyle
    O \left (w(r)k/2^k + \sum_{u \in \Phi_k}|E'_u|\log b \right).
\end{eqnarray*}

Statement 2.
 We partition $\Pi$ as follows.
 Let $\Pi'$ be the set of nodes in $\Pi$ with weight greater
 than $b^3$.
  For any $0 \leq k \leq 20$, let
$
     \Pi_k = \{ u \mid u \in \Pi \mbox{ and }
             b^{\frac{k}{7}} < w(u) \le b^{\frac{k+1}{7}} \}.
$
 Obviously, $\Pi = \Pi' \cup \Pi_0 \cup \cdots \Pi_{20}$.
 
Since $\secW(u) \le b^3$ and $w(u) > b^3$ for all nodes $u$ in $\Pi'$,
 $\Pi'$ has critical degree one.
By Lemma~\ref{lem-tradeoff}(\ref{item-heavy}), we can compute $\mwm(G_u)$ for
all nodes in $\Pi'$ using $O(\sqrt{b}w(r) + \sum_{u \in \Pi'} |E_u|)$
time.

Each  $\Pi_k$ is handled as follows.
 For every node $u \in \Pi_k$, $u$ has at most $b^{\frac{1}{7}}$
 children with weight at least $b^{\frac{k}{7}}$; otherwise
 $w(u) > b^{\frac{k+1}{7}}$ and $u\not\in\Pi_k$.
 Thus, $\Pi_k$ has critical degree $b^{\frac{1}{7}}$.
 By Lemma~\ref{lem-tradeoff}(\ref{item-heavy}), we can compute
 $\mwm(G_u)$ for all $u \in \Pi_k$ in
   $O((\sqrt{b} + b^{\frac{3}{7}}\log b )w(r) + \sum_{u \in \Pi_k} |E_u|)$
   $=$ $O(\sqrt{b}w(r) + \sum_{u \in \Pi_k}|E_u|)$ time.
  In summary, we can compute $\mwm(G_u)$ for all $u \in \Pi$
 in $O(\sqrt{b}w(r) + \sum_{u \in \Pi} |E_u|)$ time.
\end{proof}

\begin{theorem} \label{thm-hierarchical-matching}
We can compute $\mwm(G_u)$
 for all nodes $u \in T$ in
 $O(\sqrt{b}w(r) + \toe)$ time.
\end{theorem}
\begin{proof}
  It follows from Lemma~\ref{lem:phi} and
the fact that
$T = \Phi \cup \Pi$ and $\sum_{u \in T} |E_u| = \toe$.
\end{proof}

\section{Computing maximum agreement subtrees}
\label{sec:mast}

By generalizing  the work of Cole
{\em et al.}~\cite{Cole:2000:AMA}
on binary evolutionary trees, we can
easily derive an algorithm to compute
a maximum agreement subtree of two labeled trees.
There is, however, a bottleneck of computing the
maximum weight
matchings of a large number of bipartite graphs
with nonconstant degrees.
By using our result on the hierarchical
bipartite matchings,
we can eliminate this bottleneck and obtain
the fastest known $\rmast$ algorithm.
Section~\ref{sec:basic_concept} introduces
basics of labeled trees.
Section~\ref{sec:matching} uses our results
on the hierarchical bipartite matchings to
remove the bottleneck in our $\rmast$ algorithm.
Section~\ref{sec:mast_alg} details
 our $\rmast$ algorithm  and analyzes its time complexity.
Section~\ref{sec:mam} discusses the generalization
 of the work of Cole {\em et al.}~\cite{Cole:2000:AMA}.

Throughout this section,
 $T_1$ and $T_2$ denote two labeled tress with $n$ nodes
 and of degree $d \ge 2$.   Let $\M_{T_1, T_2} =
 \sum_{u \in T_1} \sum_{v \in T_2} \delta(u, v)$
 where $\delta(u, v) = 1$ if nodes $u$ and $v$
 are labeled with the same symbol, and $0$ otherwise.
 Also, let $\M$ denote $\M_{T_1,T_2}$.
 
\subsection{Basics} \label{sec:basic_concept}
\begin{figure}
 \begin{center}
 \begin{picture}(0,0)%
 \epsfig{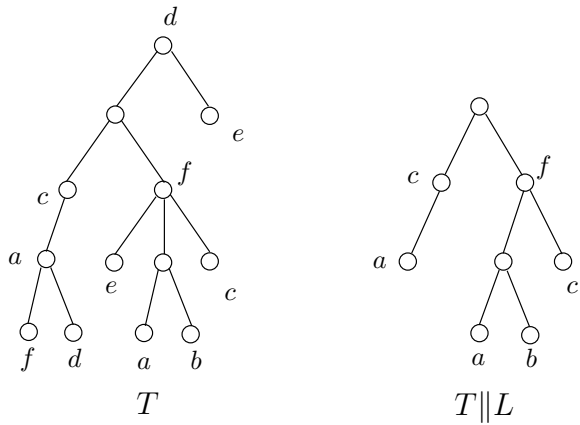}%
 \end{picture}%
 \setlength{\unitlength}{3947sp}%
 \begingroup\makeatletter\ifx\SetFigFont\undefined%
 \gdef\SetFigFont#1#2#3#4#5{%
   \reset@font\fontsize{#1}{#2pt}%
   \fontfamily{#3}\fontseries{#4}\fontshape{#5}%
   \selectfont}%
 \fi\endgroup%
 \begin{picture}(3715,2586)(118,-2137)
 \put(1058,-2132){\makebox(0,0)[b]{\smash{\SetFigFont{12}{14.4}{\familydefault}{\mddefault}{\updefault}$T$}}}
 \put(3173,-2137){\makebox(0,0)[b]{\smash{\SetFigFont{12}{14.4}{\familydefault}{\mddefault}{\updefault}$T \| L$}}}
 \put(1035,-1846){\makebox(0,0)[b]{\smash{\SetFigFont{10}{12.0}{\familydefault}{\mddefault}{\updefault}$a$}}}
 \put(1364,-1845){\makebox(0,0)[b]{\smash{\SetFigFont{10}{12.0}{\familydefault}{\mddefault}{\updefault}$b$}}}
 \put(1628,-412){\makebox(0,0)[b]{\smash{\SetFigFont{10}{12.0}{\familydefault}{\mddefault}{\updefault}$e$}}}
 \put(398,-810){\makebox(0,0)[b]{\smash{\SetFigFont{10}{12.0}{\familydefault}{\mddefault}{\updefault}$c$}}}
 \put(305,-1831){\makebox(0,0)[b]{\smash{\SetFigFont{10}{12.0}{\familydefault}{\mddefault}{\updefault}$f$}}}
 \put(599,-1835){\makebox(0,0)[b]{\smash{\SetFigFont{10}{12.0}{\familydefault}{\mddefault}{\updefault}$d$}}}
 \put(1576,-1411){\makebox(0,0)[b]{\smash{\SetFigFont{10}{12.0}{\familydefault}{\mddefault}{\updefault}$c$}}}
 \put(1201,314){\makebox(0,0)[b]{\smash{\SetFigFont{10}{12.0}{\familydefault}{\mddefault}{\updefault}$d$}}}
 \put(226,-1186){\makebox(0,0)[b]{\smash{\SetFigFont{10}{12.0}{\familydefault}{\mddefault}{\updefault}$a$}}}
 \put(830,-1372){\makebox(0,0)[b]{\smash{\SetFigFont{10}{12.0}{\familydefault}{\mddefault}{\updefault}$e$}}}
 \put(1291,-661){\makebox(0,0)[b]{\smash{\SetFigFont{10}{12.0}{\familydefault}{\mddefault}{\updefault}$f$}}}
 \put(2723,-676){\makebox(0,0)[b]{\smash{\SetFigFont{10}{12.0}{\familydefault}{\mddefault}{\updefault}$c$}}}
 \put(2516,-1212){\makebox(0,0)[b]{\smash{\SetFigFont{10}{12.0}{\familydefault}{\mddefault}{\updefault}$a$}}}
 \put(3135,-1822){\makebox(0,0)[b]{\smash{\SetFigFont{10}{12.0}{\familydefault}{\mddefault}{\updefault}$a$}}}
 \put(3469,-1836){\makebox(0,0)[b]{\smash{\SetFigFont{10}{12.0}{\familydefault}{\mddefault}{\updefault}$b$}}}
 \put(3726,-1372){\makebox(0,0)[b]{\smash{\SetFigFont{10}{12.0}{\familydefault}{\mddefault}{\updefault}$c$}}}
 \put(3545,-623){\makebox(0,0)[b]{\smash{\SetFigFont{10}{12.0}{\familydefault}{\mddefault}{\updefault}$f$}}}
 \end{picture}
 \caption{The restricted subtree $T\|L$ with $L = \{a , b, c\}$.}
 \label{fig-order-restrict}
 \end{center}
 \end{figure}

 For a rooted tree $T$ and
 any node $u$ of $T$,
 let $T^u$ denote the subtree of $T$ that is rooted at $u$.
 For any set $L$ of symbols,
 the {\it restricted subtree} of $T$ with respect to $L$, denoted by $T\|L$,
 is the subtree of $T$ (1) whose nodes are the nodes with labels from  $L$
 and the least common ancestors of any two nodes with labels from $L$
 and (2) whose edges preserve the ancestor-descendant relationship of $T$.
 Figure~\ref{fig-order-restrict} gives an example.
 Note that $T\|L$ may contain nodes
 with labels outside $L$.
 For any labeled tree $T'$,
 let $T\|T'$ denote the restricted subtree of $T$ with respect to
 the set of symbols used in $T'$.

\begin{figure}
 \begin{center}
 \epsfig{file=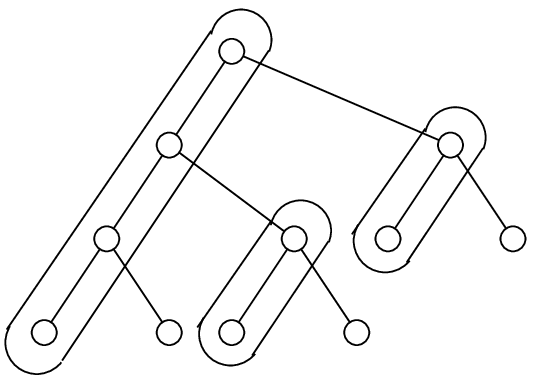, height=1in}
 \end{center}
 \caption{A centroid path decomposition of a rooted tree.}
 \label{fig-centroid-path}
 \end{figure}

 A {\it centroid path decomposition} \cite{Cole:2000:AMA}
 of a rooted tree $T$
 is a partition of its nodes into
 disjoint paths as follows.
 For each internal node $u$ in $T$,
 let $C(u)$ denote the set of children of $u$.
 Among the children of $u$, one is chosen as  the {\em heavy} child,
 denoted by $\cntr(u)$, if the subtree of $T$ rooted at $\cntr(u)$
 contains the largest number of nodes;
 the other children of $u$ are {\em side} children.
 We call the edge from $u$ to its heavy child
 a {\it heavy} edge.
 A {\it centroid path} is a maximal path formed by heavy edges;
 the {\it root} centroid path is the centroid path
 that contains the {\it root}\/ of $T$.
 See Figure~\ref{fig-centroid-path}
  for an example.

 \newcommand{\rt}[1]{r({\scriptstyle #1})}
 \newcommand{\rtt}[1]{r{\scriptscriptstyle (#1)}}

 Let $\cpdecom(T)$ denote the set of the centroid paths of $T$. Note
 that $\cpdecom(T)$
 can be constructed in $O(|T|)$ time.
 For every $P \in \cpdecom(T)$, the root of $P$, denoted $r(P)$,
 refers to
 the node on $P$ that is the closest to the root of $T$, and
 $\cpattach(P)$ denotes the set of the side children of the nodes on $P$.
 For any node $u$ on $P$,
 a subtree rooted at some side child of $u$ is called
 a {\it side tree} of $u$, as well as a {\it side tree} of $P$.
 Let $\sideT(P)$ be the set of side trees of $P$.
 Note that for every $R \in \sideT(P)$, $|R| \leq |T^{\rtt{P}}|/2$.

 The following lemma states two useful
 properties of the centroid path decomposition.

 \begin{lemma}\label{lemma-1}\
 Let $T_1$ and $T_2$ be two labeled trees.
 
 \begin{enumerate}
 \item   $\sum_{P \in \cpdecom(T_1)}   \M_{T_1^{^\rtt{P}},T_2}$
       $\leq \M_{T_1,T_2} \log n$.
 \item
 $\sum_{P \in \cpdecom(T_1)}  \sqrt{ \min( d, |T_1^{\rtt{P}}|) }
 \M_{T_1^{\rtt{P}},T_2}$
 $\leq \sqrt{d} \M_{T_1,T_2} \log \frac{2n}{d}$.
 \end{enumerate}
 \end{lemma}

 \begin{proof}
The two statements are proved as follows.

 Statement 1.
 A centroid path $P$ is {\em attached} to another
 centroid path $P'$ if the root of $P$ is the child of a node on $P'$.
 We define the {\em level}\/ of a centroid path as follows.  The root
 centroid path has level zero.  A centroid path has level $i$ if it is
 attached to some centroid path with level $i-1$.  Note that any
 subtree attached to a centroid path with level $i$ has size at most
 $n/2^{i+1}$.  Thus, there are at most $\log n$ different levels.
 Moreover, subtrees attached to centroid paths with the same level are
 all disjoint.

 For any $0 \leq i < \log n$, denote by $D_i$ the set of
 all centroid paths in ${\cal D}(T_1)$ with level $i$.  Then
 $\sum_{P \in \cpdecom(T_1)}   \M_{T_1^{\rtt{P}},T_2}$
 $ = \sum_{0 \leq i < \log n}
 \sum_{P \in D_i}   \M_{T_1^{\rtt{P}},T_2}$
 $ \leq \log n  \sum_{P \in D_i}   \M_{T_1^{\rtt{P}},T_2}$
       $\leq \M_{T_1,T_2} \log n$.

 Statement 2. We divide the centroid paths into 2 groups.
 We first consider the centroid paths on level $i$ where
 $0 \leq i <  \log \frac{2n}{d}$.
 For any such $i$,
\[
\sum_{P \in D_i}\sqrt{\min\{d,|T_1^{\rtt{P}}|\}}\M_{T_1^{\rtt{P}},T_2}
\leq \sum_{P \in D_i}\sqrt{d}\M_{T_1^{\rtt{P}},T_2}
\leq \sqrt{d}\M_{T_1,T_2}.
\]
Thus,
  $\sum_{0 \leq i < \log \frac{2n}{d}}
   \sum_{P\in D_i}\sqrt{\min\{d,|T_1^{\rtt{P}}|\}}\M_{T_1^{\rtt{P}},T_2}
   \leq \sqrt{d}\M_{T_1,T_2} \log \frac{2n}{d}$.

Next, we consider the centroid paths on level
$\log \frac{2n}{d} + i$ where $i \geq 0$.
Note that for a path $P$ on level $\log \frac{2n}{d} + i$,
$|T_1^{\rtt{P}}| \leq d/2^{i+1}$.
Thus,
$\sum_{i \geq 0} \sum_{P \in D_{\log\frac{2n}{d}+i}}
              \sqrt{\min\{d,|T_1^{\rtt{P}}|\}}\M_{T_1^{\rtt{P}},T_2}$
is at most
\[
\sum_{i \geq 0} \sum_{P \in D_{\log\frac{2n}{d}+i}}
\sqrt{d/2^{i+1}} \M_{T_1^{\rtt{P}},T_2}
\leq
\sum_{i \geq 0}
\sqrt{d/2^{i+1}} \M_{T_1,T_2}
= O(\sqrt{d}\M_{T_1,T_2}).
\]

\end{proof}

The following notion captures which pairs of nodes
of two labeled trees $T_1$ and $T_2$ are important.
Consider any centroid paths $P \in \cpdecom(T_1)$
and $Q \in \cpdecom(T_2)$.
 For any node $x \in P$ or $Q$, let
 ${\cal L}(x)$ be the set of symbols
 labeling $x$ and the nodes
 in the side trees of $x$.
 Let $\inp(P,Q)$ be the set of node pairs $(u,v) \in P \times Q$
 with ${\cal L}(u) \cap {\cal L}(v) \neq \emptyset$.
 Let  $\inp(P,T_2) = \bigcup_{Q \in \cpdecom(T_2)} \inp(P,Q)$.

\subsection{Matchings} \label{sec:matching}
As explained later in \S\ref{sec:mast_alg},
  we can easily generalize the dynamic programming approach
  in \cite{Cole:2000:AMA} to compute $\rmast(T_1, T_2)$
  for any two labeled trees $T_1$ and $T_2$, but
there is a bottleneck of computing the maximum weight matchings of
a large number of bipartite graphs with nonconstant degrees.
This section uses our results on
hierarchical bipartite matchings to remove this bottleneck.

First of all, we identify the bipartite graphs
for which maximum weight matchings are required.
For any nodes $u \in T_1$ and $v \in T_2$,
define $G_{uv}$ as the weighted bipartite graph between
  $C(u)$ and $C(v)$ where edge $(x,y)$ has weight
  $\rmast(T_1^x, T_2^y)$.
Furthermore, define
$H_{uv}$ as the graph constructed from $G_{uv}$ by removing
all the zero-weight edges and all the edges adjacent to
the heavy child of $u$ or $v$.
   Note that the total edge weight
   of $H_{uv}$ can be significantly smaller than that of $G_{uv}$.
   Yet by Lemma~\ref{lem-recover}, we can recover $\mwm(G_{uv})$ from
   $\mwm(H_{uv})$ efficiently.

\newcommand{\tnoe}{\mbox{\rm tnoe}}
\newcommand{\TM}{{\cal T\!M}}

This section shows that for any centroid path $P \in \cpdecom(T_1)$,
we can efficiently compute $\mwm(H_{uv})$ for all $(u,v) \in \inp(P,T_2)$.
More precisely,
let $\TM_P$ denote the required time;
the key result of this section
is that
$\sum_{P \in \cpdecom(T_1)}\TM_P \leq$ $\sqrt{d} \M \log \frac{2n}{d}$
(see Lemma~\ref{Mstar}).

To derive an upper bound on $\TM_P$, we need an
estimate of the number of edges
in the graphs $H_{uv}$ for all $(u,v) \in \inp(P,T_2)$.
For any centroid path $P \in \cpdecom(T_1)$, let
$\tnoe(P)$ be the total number of edges in
the graphs $H_{uv}$ for all $(u,v) \in \inp(P,T_2)$.
Furthermore, let $\tnoe =$
$\sum_{P \in \cpdecom(T_1)} \tnoe(P)$.

\begin{lemma}\label{Huv}\
\begin{enumerate}
\item
For any
$P \in \cpdecom(T_1)$,
$\tnoe(P) =$
$O\left( \sum_{ w \in \cpattach(P)} |T_2 \| T_1^w|
\log \frac{2 | T_2 \| T_1^{\rtt{P}}|}{|T_2 \|  T_1^w |}
\right)$.
\item $\tnoe = O(\M \log n)$.
\end{enumerate}
\end{lemma}
\begin{proof}
The two statements are proved as follows.

Statement 1.
Let $r$ be the root of $P$.  By definition,
every edge in $H_{uv}$ for any
$(u, v) \in \inp(P, T_2)$
corresponds to a pair of side trees $(\Upsilon, \Pi)$
where $\Upsilon \in \sideT(P)$ and $\Pi \in \sideT(Q)$ for
some $Q \in \cpdecom(T_2 \| T_1^r)$
such that $\Upsilon$ and $\Pi$ contain some common labels.
We call $(\Upsilon, \Pi)$ an {\it intersecting} side tree pair.
Thus,
$\tnoe(P)$ is at most the total number of intersecting side tree
pairs in $\sideT(P) \times
\bigcup \{\sideT(Q) \mid Q \in  \cpdecom(T_2 \| T_1^r)\}$.

To simplify our discussion, let $R=T_2\|T_1^r$
and $\sideT(R)=\bigcup \{\sideT(Q) \mid Q \in  \cpdecom(R)\}$.
Consider any node $w \in {\cal A}(P)$.  $T_1^w$ is a side tree in $\sideT(P)$.
Let $R_w$ be $T_2\|T_1^w$.
Note that each path in $R_w$ starting
from a node $x$ to its descendant $y$
corresponds to a simple path $Q_{xy}$ in $R$ from $x$ to $y$.
Let $1^{st}(x,y)$ be the node on $Q_{xy}$ which is the child of $x$.
By the definition of side trees, among all the side trees in $\sideT(R)$,
at most $\log |R^{1^{st}(x,y)}| + 1$
have roots on $Q_{xy}$.

\newcommand{\summ}{{\sc sum}}

For all side trees $R^v \in \sideT(R)$,
$(T_1^w, R^v)$ is an intersecting side tree pair if and only if
either
(1) $v$ is a node on the path from the root of $R$
to the root $R_w$; or
(2) $v$ is a node on some path $Q_{xy}$ on $R$
where $(x, y)$ is an edge in $R_w$.
The number of side trees $R^v \in \sideT(R)$ in case (1)
is less than $\log |R|$.
The number of side trees $R^v \in \sideT(R)$ in case (2)
is less than
$\sum_{(x, y) \in R_w}
\left(\log |R^{1^{st}(x,y)}| + 1 \right)$.
Let \summ$(R_w)$ denote $\sum_{(x, y) \in R_w} \log |R^{1^{st}(x,y)}|$.
Below we prove \summ$(R_w)$
= $O\left(|R_w|  \log \frac{2|R|}{|R_w|} \right)$.
In total,
$\tnoe(P) =$
$O\left( \sum_{w \in {\cal A}(P)} \left\{
|R_w| \log \frac{2|R|}{|R_w|} \right\}\right)$,
as claimed in this statement.

It remains to prove
\summ$(R_w) = O\left(|R_w|
 \log \frac{2|R|}{|R_w|}\right)$.
For any leaf $y$ of $R$, let $p(y)$ be the maximal path in $R$
ending at $y$
such that every node on $p(y)$ has at most one child;
denote $r_p(y)$ as the root of $p(y)$.
Let $Z_{R_w} = \{ p(y) \mid y$ is a leaf of $R_w \}$.
As $\{ R^{1^{st}(r_p(y), y)} \mid p(y) \in Z_{R_w}\}$
is a set of disjoint subtrees of $R$,
$\sum_{p(y) \in Z_{R_w}} |R^{1^{st}(r_p(y), y)}| \leq |R|$.
Note that $|Z_{R_w}| \leq |R_w|$.
Thus,
$\sum_{p(y) \in Z_{R_w}} \log |R^{1^{st}(r_p(y), y)}| \leq$
$|R_w| \log \frac{2 |R|}{|R_w|}$.\footnote{This follows from the fact that
for any sequence of positive numbers $a_1, a_2, \ldots, a_k$
with the sum equal to $s$,
$\sum_{i=1}^{k} \log a_i \leq \ell \log \frac{2s}{\ell}$,
where $k \leq \ell \leq s$.
}
Let $\hat{R}_w$ be the tree obtained by removing all the paths in
$Z_{R_w}$.
We have \summ$(R_w) = |R_w| \log \frac{2|R|}{|R_w|} +$
\summ$(\hat{R}_w)$.
Note that $\hat{R}_w$ contains at most half the leaves of $R_w$.
Hence,
\summ$(R_w) =
O\left(|R_w| \log \frac{2|R|}{|R_w|}\right)$.

Statement 2.
By Statement 1,
$\tnoe$ is in the order of
\begin{eqnarray*}
& & \lefteqn{
\sum_{P \in \cpdecom(T_1)}
\left[ \sum_{w \in \cpattach(P)} |T_2 \| T_1^w|
\log \frac{ 2 | T_2 \| T_1^{\rtt{P}} |}{  |T_2 \|  T_1^w |}
\right]
} \\
& \leq &
\sum_{P \in \cpdecom(T_1)}
\left[ \sum_{w \in \cpattach(P)} \M_{T_1^w,T_2}
\log \frac{ 2 | T_2 \| T_1^{\rtt{P}} |}{  \log |T_2 \|  T_1^w |}
\right]\\
& = &
\sum_{P \in \cpdecom(T_1)}
\left[
\sum_{w \in \cpattach(P)} \M_{T_1^w,T_2} (1 +
\log |T_2\|T_1^{\rtt{P}}|  -\log {|T_2\|T_1^{w}|})
\right] \\
& \leq &
\sum_{P \in \cpdecom(T_1)}
\left[
\M_{T_1^{\rtt{P}},T_2} +
\M_{T_1^{\rtt{P}},T_2} \log |T_2\|T_1^{\rtt{P}}|  -
\sum_{w \in \cpattach(P)}
\M_{T_1^w,T_2} \log {|T_2\|T_1^{w}|}
\right] \\
& \leq &
\sum_{P \in \cpdecom(T_1)} \M_{T_1^{\rtt{P}},T_2} +
\sum_{P \in \cpdecom(T_1)}
\left[
\M_{T_1^{\rtt{P}},T_2} \log |T_2\|T_1^{\rtt{P}}|  -
\sum_{w \in \cpattach(P)}
\M_{T_1^w,T_2} \log {|T_2\|T_1^{w}|}
\right] \\
& = &
\sum_{P \in \cpdecom(T_1)} \M_{T_1^{\rtt{P}},T_2} +
\M_{T_1^{r_o},T_2} \log {|T_2\|T_1^{r_o}|}
\mbox{, where $r_o$ is the root of $T_1$}\\
& \le &
\sum_{P \in \cpdecom(T_1)} \M_{T_1^{\rtt{P}},T_2} +
\M_{T_1,T_2}\log {|T_2|} \\
& \le &
\M_{T_1,T_2}\log {|T_2|} + \M_{T_1,T_2}\log {|T_2|}
\hspace{.5in}\mbox{ by Lemma~\ref{lemma-1}(1)}
\\
& = & 2 \M\log n.
\end{eqnarray*}
\end{proof}

We proceed to detail the computing of $\mwm(H_{uv})$
for all $(u,v) \in \bigcup_{P \in \cpdecom(T_1)} \inp(P,T_2)$.
A bipartite graph is {\em nontrivial}\/
if both node sets have at least two nodes.
Computing $\mwm(H_{uv})$ for all trivial $H_{uv}$
takes only linear time, i.e., $O(\tnoe)$ = $O(\M\log n)$ time.
Thus, we focus on those nontrivial $H_{uv}$.

Consider any centroid path $P$ in $\cpdecom(T_1)$
and fix a node $u$ of $P$.
Let ${\cal H}_u$  be the set of all nontrivial graphs $H_{uv}$
where $(u,v) \in {\inp}(P,T_2)$.
Let $\TM_u$  be the time for finding
$\mwm(H_{uv})$ for all the graphs in ${\cal H}_u$.
Let $\tnoe(u)$ be the number of edges of
all the graphs in ${\cal H}_u$.
In the next lemma,
we first derive an upper bound of $\TM_u$,
and then we show
$\sum_{P \in \cpdecom(T_1)}\TM_P$ =
$O(\sqrt{d}\M \log\frac{2n}{d})$.

{\samepage
\begin{lemma}\label{Mstar}\
\begin{enumerate}
\item
$\TM_u = O(\sqrt{\min(d,|T_1^u|)}\M_{S_u,T_2} + \tnoe(u))$,
where $S_u$ is the set of side trees of $u$ in $T_1$ and
$\M_{S_u,T_2} = \sum_{\Upsilon \in S_u} \M_{\Upsilon, T_2}$.
\item
$\sum_{P \in \cpdecom(T_1)}\TM_P$ =
$O(\sqrt{d}\M \log\frac{2n}{d})$.
\end{enumerate}
\end{lemma}
}

\begin{proof}
The two statements are proved as follows.

Statement 1.
Let $B_u$ be the set of labels used in the side trees in $S_u$.
First, we show that for all $v \in T_2 \| B_u$,
$\mwm(H_{uv})$ can be computed in $O\Big(\sqrt{\min(d,|T_1^u|)}\M_{S_u,T_2} +
\tnoe(u)\Big)$ time.
Second,
we recover $\mwm(H_{uv})$ for all nontrivial
$H_{uv}$ where $(u,v) \in \inp(P,T_2)$ in $O(\M_{S_u,T_2})$ time.
Then this statement follows.

To compute $\mwm(H_{uv})$ for all $v \in T_2 \| B_u$,
we apply the hierarchical bipartite matching algorithm of
\S\ref{sec:hi-matching}.
Let $T = T_2 \| B_u$.
For every node $v \in T$, we
associate with $v$
the bipartite graph $H_{uv}$ and let $w(v) = \M_{S_u, T_2^v}$.
Observe that $w(v) = \M_{S_u, T_2^v} \geq \sum_{x \in C(v)} \M_{S_u, T_2^x}
= \sum_{x \in C(v)} w(x)$.
In addition, for every node $x \in C(v)$, the total weight of all
the edges incident to $x$ in $H_{uv}$ is at most $w(x) = \M_{S_u, T_2^x}$.
Hence, $T$ and the associated bipartite graphs $H_{uv}$ satisfy
the conditions for the hierarchical bipartite matching problem.
For the time complexity,
note that, for every $v \in T$, the two node sets of $H_{uv}$
have size bounded by $d$ and $\min( d, |T_1^u|)$, respectively.
Thus, by Theorem~\ref{thm-hierarchical-matching},
we can find $\mwm(H_{uv})$ for
all nodes $v$ of $T = T_2 \| B_u$ in
$O(\sqrt{\min(d,|T_1^u|)}\M_{S_u,T_2} + \tnoe(u))$ time.

Next, we show how to recover $\mwm(H_{uv})$ for all $v \in L$,
where $L$ denotes the set of nodes $v$ of $T_2$ such that
$(u,v) \in \inp(P,T_2)$ and $H_{uv}$ is nontrivial.
Note that every node $x$ of $T_2 \| B_u$ is also a node in $T_2$ and
every edge $(x, y)$ of $T_2 \| B_u$ corresponds to a path in $T_2$.
Also observe that
every node $v \in L$ must be
a node in $T_2 \| B_u$;
otherwise,
$v$ lies on a path corresponding to
an edge $(x, y)$ of $T_2 \|B_u$, and
$H_{uv}$ contains a singleton node set
and is trivial because $v \not\in L$.
Therefore, we can compute $\mwm(H_{uv})$
for all $v \in L$ by traversing
$T_2 \| B_u$ once using $O(|(T_2\|B_u)|) = O(\M_{S_u,T_2})$ time.

Statement 2.
By Statement 1,
\begin{eqnarray*}
\sum_{P \in \cpdecom(T_1)}\TM_P & = &
     O\left( \sum_{P \in \cpdecom(T_1)}\sum_{u \in P} \TM_u \right)\\
& = &       O\left( \sum_{P \in \cpdecom(T_1)}
  \sum_{u \in P} ( \sqrt{\min(d,|T_1^u|)}\M_{S_u,T_2} + \tnoe(u)) \right)\\
& = &       O\left(\tnoe + \sum_{P \in \cpdecom(T_1)}
  \sum_{u \in P}  \sqrt{\min(d,|T_1^u|)}\M_{S_u,T_2} \right)\\
& = & O\left(\tnoe +
\sum_{P \in \cpdecom(T_1)}\sqrt{\min(d,|T_1^{\rtt{P}}|)}\M_{T_1^{\rtt{P}},T_2}
\right)\\
& = & O\left(\tnoe +
\sqrt{d} \M_{T_1,T_2} \log \frac{2n}{d}\right)
\mbox{\hspace*{.3in}by Lemma~\ref{lemma-1}}\\
& = & O\left(\sqrt{d}\M \log\frac{2n}{d}\right)
\mbox{\hspace*{.3in}by Lemma~\ref{Huv}}
\end{eqnarray*}
\end{proof}

\subsection{The MAST algorithm}\label{sec:mast_alg}
Our algorithm is based on the following
recurrence, which generalizes the one given in \cite{Farach:1997:SDP}
to handle labeled trees.
 
 \newcommand{\tw}[1]{\|#1\|}
 \begin{equation} \label{basic-equation}
  \rmast(T_1^u,T_2^v) = \max \left \{
      \begin{array}{l}
        \max\{\rmast(T_1^u,T_2^x) \mid x \in C(v)\}, \\
        \max\{\rmast(T_1^x,T_2^v) \mid x \in C(u)\}, \\
        \tw{\mwm(G_{uv})} \mbox{ if $u$ and $v$ are both unlabeled}, \\
        \tw{\mwm(G_{uv})}+1
        \mbox{ for $u, v$ labeled with the same symbol,} \\
      \end{array}
    \right .
 \end{equation}
where $\tw{\mwm(G_{uv})}$ denotes the total weight of the
matching.

 Equation~(\ref{basic-equation}) suggests a bottom-up
 dynamic programming approach to computing \rmast$(T_1,T_2)$.
 The following lemma
 generalizes the
 technique of Cole {\em et~al.}~\cite{Cole:2000:AMA} for speeding up the
 dynamic programming.
 Basically, it states that the time complexity is dominated by the
 time for finding maximum weight matchings of some graphs $H_{uv}$.

 \begin{lemma} \label{basic-lemma}
Let $P \in \cpdecom(T_1)$ be a centroid path and $r = r(P)$.
 Given
 the values $\rmast(T_1^u, (T_2\|T_1^u)^v)$
 for all nodes $u \in \cpattach(P)$
 and $v \in T_2\|T_1^u$,
 we can compute $\rmast(T_1^{r}, (T_2\|T_1^r)^v)$ for
 all $v \in T_2\|T_1^{r}$ in
 $O\left(
   \left(\gamma(T_1^r) -  \sum_{u \in \cpattach(P)} \gamma(T_1^u) + \M_{T_1^r, T_2}\right)
 \log d  + \TM_P \right)$ time,
 where $\gamma(R)$ denotes
 $\M_{R,T_2} \log |(T_2\|R)|$.
 \end{lemma}

Cole {\em et~al.}~\cite{Cole:2000:AMA} proved Lemma~\ref{basic-lemma}
for the special case where
$T_1$ and $T_2$ are binary evolutionary trees.
For a better flow of  discussion,
we postpone the proof of Lemma~\ref{basic-lemma} to \S\ref{sec:mam}.
Here, Lemma~\ref{basic-lemma} immediately suggests that
$\rmast(T_1,T_2)$ can be computed in a bottom-up fashion as follows:
\begin{itemize}
\item
Step 1.
  Let $\prec$ denote the ordering on $\cpdecom(T)$ where
  $P_1 \prec P_2$ if the root of $P_1$ is a descendant of  the root $P_2$.

\item
Step 2.
 For every $P \in \cpdecom(T_1)$ in increasing order
         according to $\prec$,
         let $r$ denote the root of $P$;
         apply Lemma~\ref{basic-lemma} to find
         $(T_1^{{r}}, (T_2\|T_1^{{r}})^v)$ for every node
         $v \in T_2\|T_1^r$.
\end{itemize}

 The above algorithm at the end computes
  $\rmast(T_1^{r(P_o)}, (T_2\|T_1^{r(P_o)}))$, where $P_o$ is
  root centroid path of $T_1$.  Since
  $r(P_o)$ is also the root of $T$, we have
  $T_1^{r(P_o)} = T_1$ and
  $\rmast(T_1^{r(P_o)}, (T_2\|T_1^{r(P_o)})) = \rmast(T_1, T_2)$.
As stated in the following lemma, the running time is
dominated by the time for computing the maximum weight matchings,
i.e., $\sum_{P \in \cpdecom(T_1)} \TM_P$.

\begin{lemma} \label{lem-time-complexity}
We can compute $\rmast(T_1,T_2)$ in
$O( \M \log n \log d + \sum_{P \in \cpdecom(T_1)} \TM_P)$
time, where $\M = \M_{T_1,T_2}$.
\end{lemma}
\begin{proof}
To derive the time for computing $\rmast(T_1,T_2)$, we simply
sum the time bound stated in Lemma~\ref{basic-lemma}
over all centroid paths of $T_1$.
Observe that
\begin{eqnarray*}
\sum_{P \in \cpdecom(T_1)}
\left[
  \gamma(T_1^{\rtt{P}}) -  \sum_{u \in \cpattach(P)} \gamma(T_1^u)
\right]
& = &
\gamma(T_1^{r_o})  \mbox{, where $r_o$ is the root of
of $T_1$}\\
& = &
\M_{T_1, T_2} \log {|T_2\|T_1|}  \\
& = &
\M_{T_1,T_2}\log {|T_2|} \\
& = & \M\log n.
\end{eqnarray*}
Thus, we can compute $\rmast(T_1,T_2)$ in
$O(\M\log n \log d + \sum_{P \in \cpdecom(T_1)}\TM_P$ $ +
\sum_{P \in \cpdecom(T_1)}\M_{T_1^{\rtt{P}},T_2}\log d)$ time.
By Lemma~\ref{lemma-1},
$\sum_{P \in \cpdecom(T_1)}\M_{T_1^{\rtt{P}},T_2} \le \M\log n$.
Thus, this lemma follows.
\end{proof}

\begin{theorem} \label{thm:mast}
$\rmast(T_1,T_2)$ can be computed
in $O(\sqrt{d}\M_{T_1,T_2} \log \frac{2n}{d})$ time.
\end{theorem}
\begin{proof}
By Lemma~\ref{Mstar},
$\sum_{P \in \cpdecom(T_1)}\TM_P$ =
$O(\sqrt{d}\M \log\frac{2n}{d})$.
Thus, by Lemma~\ref{lem-time-complexity},
\rmast$(T_1,T_2)$ can be computed in
$O(\M (\log n\log d + \sqrt{d} \log\frac{2n}{d}))$ time.
Since $\log n \log d \leq \sqrt{d} \log\frac{2n}{d}$,
this theorem follows.
\end{proof}

\subsection{Proof of Lemma~\ref{basic-lemma}}\label{sec:mam}
This section provides the details for adapting the techniques
of Cole {\em et al.}~\cite{Cole:2000:AMA} to prove Lemma~\ref{basic-lemma}.
Consider any centroid path  $P \in {\cal D}(T_1)$. Let
$r$ be the root of $P$.  Lemma~\ref{basic-lemma} states that
if we are given, for every $u \in \cpattach(P)$,
\[\textstyle
 \rmast(T_1^u, (T_2 \| T_1 ^u)^v) \mbox{ for all }
        v \in T_2 \| T_1^u,
\]
then we can compute
\begin{equation} \label{require_value} \textstyle
    \rmast(T_1^r, (T_2 \| T_1^r)^v) \mbox{ for all } v \in T_2 \| T_1^r
\end{equation}
in $ O\left(
   (\gamma(T_1^r) -  \sum_{u \in \cpattach(P)} \gamma(T_1^u) +
     \M_{T_1^r, T_2} )
 \log d  + \TM_P \right)$ time.

The centroid paths in $\cpdecom(T_2\|T_1^r)$ partition
the set of nodes of $T_2\|T_1^r$ and define an ordering
on the nodes of $T_2\|T_1^r$.
Precisely, the set of  values in
 Equation~(\ref{require_value}) are partitioned into the
 following sets
\[
\textstyle
 \{ \rmast(T_1^r, (T_2 \| T_1^r)^v) \mid v \in Q \}
 \mbox{  where } Q \in \cpdecom(T_2 \| T_1^r).
\]
We focus on computing
  $\{ \rmast(T_1^r, (T_2 \| T_1^r)^v) \mid v \in Q \}$ for
each $Q \in \cpdecom(T_2\|T_1^r)$.
Cole {\em et~al.}~\cite{Cole:2000:AMA}
dealt with the special case where
$T_1$ and $T_2$ are binary evolutionary trees.
They introduced the {\em maximum agreement matching} (MAM) problem
and showed that
$\{ \rmast(T_1^r, (T_2\|T_1^r)^v) \mid v \in Q\}$ can be
computed by solving the MAM problem
on some weighted bipartite multigraph.
In \cite{Przytycka:1997:SDP}, Przytycka observed that
this technique can be generalized to
evolutionary trees of arbitrary degrees;
basically, it suffices to use a more complicated bipartite multigraph.
We observe that
this can be further generalized to labeled
trees with arbitrary degrees
by adding more edges to the multigraph.

In the rest of this section, we define the maximum
agreement matching problem and
the weighted bipartite multigraph ${\cal G}_{PQ}$ for
handling labeled trees with general degrees.

\paragraph{The maximum agreement matching problem.}
Let $\mgraph = (X,Y,E)$ be a
weighted bipartite multigraph.
Suppose that $X = \{u_1, u_2, \ldots, u_p\}$,
$Y = \{v_1, v_2, \ldots, v_q\}$, and
every pair of nodes are connected by at most four edges.
Every edge is colored by either
{\em gray, green, red,} or {\em white}.
We say that edge $(u_i, v_j)$ is {\em below} edge $(u_k,v_\ell)$
if $i < k$ and $j < \ell$; and that $(u_i, v_j)$ {\em crosses}
$(u_k, v_\ell)$ if $i < k$ and $j > \ell$.
A matching of $\mgraph$ is an {\em agreement} matching
if it satisfies all the following properties:
\begin{itemize}
\item No white edge crosses another white edge.
\item There is at most one gray edge.  If a gray edge is present, it
      must be below all the white edges.
\item There are at most one pair of red and green edges.  If such
   a pair is present,
      then this pair of edges are below all white edges, and
      the red edge crosses the green edge.
\item A gray edge cannot coexist with a pair of red and green edges.
\end{itemize}
The {\em weight} of an agreement matching of $\mgraph$ is the
total weight of the edges in the matching.  A {\em maximum
agreement matching} is one with the maximum weight, and we
denote this weight as $\mam(\mgraph)$.

For any nodes $u_i \in X$
and $v_j \in Y$, let
$\mgraph(u_i, v_j)$ denote the subgraph
of $\mgraph$ induced by the nodes
$u_i, u_{i+1}, \ldots, u_p$ and $v_j, v_{j+1},\ldots, v_q$.
The maximum agreement matching problem asks for
$\mam(\mgraph(u_i,v_j))$
for all pairs of $(u_i, v_j)$
such that either
(1) $u_i = u_1$ and $v_j$ is adjacent to some edges of $\mgraph$; or
(2) $v_j = v_1$ and $u_i$ is adjacent to some edges of $\mgraph$.

\paragraph{The weighted bipartite multigraph \boldmath$\mgraph_{PQ}$.}
Roughly speaking, $\mgraph_{PQ}$ is constructed by
adding suitable colored edges between $P$ and $Q$.
Our aim is that by solving the MAM problem on
$\mgraph_{PQ}$, all the values in Equation~(\ref{require_value})
are found automatically.

First, we define a new graph
$H'_{uv}$ from $G_{uv}$ and $H_{uv}$ as follows.
$H'_{uv}$ has all the edges of $H_{uv}$, as well as
some other edges from $G_{uv}$.
Among all the edges of $G_{uv}$ adjacent
to $\cntr(u)$, we add into $H'_{uv}$ those edges $(\cntr(u), y)$ where
$y$ is adjacent to some edges of $H_{uv}$. Among the rest of
the edges adjacent to $\cntr(u)$, we add into $H'_{uv}$ the
one with the heaviest weight.  Similarly,
among all edges adjacent to $\cntr(v)$,
we choose some edges to add into $H'_{uv}$.

We are now ready to define $\mgraph_{PQ}$.
Suppose that $P = (u_1, u_2, \ldots, u_p)$ and
$Q = (v_1, v_2, \ldots, v_q)$.
There is one or more edges between nodes $u_i$ and $v_j$ if and only
if $(u_i, v_j) \in \inp(P, Q)$.  The number, color, and weight of edges
between $u_i$ and $v_j$ are determined in the three cases below.
Let $\maxsideR =\max\{\rmast(T_1^{u_i},\Gamma) \mid \Gamma$
is a side tree of $v_j \}$.
Let
$\maxsideL = \max\{\rmast(\Gamma, T_2^{v_j}) \mid \Gamma$
is a side tree of $u_i \}$.

{\it Case} 1:
$u_i$ and $v_j$ are both unlabeled.
There are a white edge, a gray edge, a green edge and a red edge
connecting $u_i$ and $v_j$, with weights $\tw{\mwm(H_{u_iv_j})}$,
$\tw{\mwm(H'_{u_iv_j})}$, $\maxsideR$, and $\maxsideL$, respectively.

{\it Case} 2:
$u_i$ and $v_j$ are labeled by the same symbol $z$.
There are a white edge and a gray edge connecting them.
The weight of the white edge is $\tw{\mwm(H_{u_iv_j})} + \mu(z)$.
The weight of the gray edge equals the maximum of
$\tw{\mwm(H'_{u_iv_j})}$
$+\mu(z)$,
$\maxsideR$, and $\maxsideL$.

{\it Case} 3:
either $u_i$ and $v_j$ are labeled by different symbols,
or only one of them is labeled.
There is only one gray edge connecting them.  Its weight equals
the larger of $\maxsideR$ and $\maxsideL$.

Note that when the input is evolutionary trees,
$\mgraph_{PQ}$ is reduced to
the multigraph defined in \cite{Przytycka:1997:SDP}, in which
most of the edges are from Case 1, and there are
edges $(u_i, v_j)$ from Cases 2 and 3 only when $u_i$ and
$v_j$ are leaves. For labeled trees, we simply add
extra edges in
Cases 2 and 3  when $u_i$ or
$v_j$ is a labeled internal node.
By construction, we have the following fact.
\begin{fact} \label{corollary-mam-rmast}
For any $u_i \in P$ and $v_j \in Q$,
$\mam(\mgraph_{PQ}(u_i,v_j)) = \rmast(T_1^{u_i}, (T_2\|T_1^r)^{v_j})$.
Thus,
solving the $\mam$ problem on $\mgraph_{PQ}$  gives
$\rmast(T_1^r, (T_2\|T_1^r)^v)$ for all $v \in Q$.
\end{fact}

Using the techniques of Cole {\it et al.}~\cite{Cole:2000:AMA,Przytycka:1997:SDP},
we can construct $\mgraph_{PQ}$
for all $Q \in \cpdecom(T_2\|T_1^r)$
and solve the corresponding MAM problems
in $O\bigg(\Big(\gamma(T_1^r)-\sum_{u \in \cpattach(P)}\gamma(T_1^u) +
          \M_{T_1^r, T_2}\Big)\log d +\TM_P\bigg)$ total time.
Therefore, Lemma~\ref{basic-lemma} follows.

\section*{Acknowledgments}
We wish to thank anonymous referees for extremely helpful suggestions.

\bibliographystyle{plain}
\bibliography{all}
\end{document}